\documentclass[runningheads]{llncs}
\usepackage{graphicx}
\usepackage{tikz}
\usepackage{comment}
\usepackage{amsmath,amssymb} % define this before the line numbering.
\usepackage{color}
\usepackage[accsupp]{axessibility}  % Improves PDF readability for those with disabilities.

\usepackage{algorithm}  
\usepackage{algorithmicx}
\usepackage[noend]{algpseudocode}  
\usepackage{booktabs}
\usepackage{multirow}
\usepackage{cite}
\usepackage{subfigure}
\usepackage[pagebackref,breaklinks,colorlinks]{hyperref}

\begin{document}
\pagestyle{headings}
\mainmatter
\def\ECCVSubNumber{3661}  % Insert your submission number here

\title{AcroFOD: An Adaptive Method for Cross-domain Few-shot Object Detection}

\begin{comment}
\titlerunning{ECCV-22 submission ID \ECCVSubNumber} 
\authorrunning{ECCV-22 submission ID \ECCVSubNumber} 
\author{Anonymous ECCV submission}
\institute{Paper ID \ECCVSubNumber}
\end{comment}
%******************

%\begin{comment}
\titlerunning{AcroFOD}
% \orcidID{0000-1111-2222-3333}
\author{Yipeng Gao\inst{1,3} \and
Lingxiao Yang\inst{1} \and
Yunmu Huang\inst{2} \and \\
Song Xie\inst{2} \and
Shiyong Li\inst{2} \and
Wei-Shi Zheng\inst{1,3,4}*}
\authorrunning{Gao et al.}
\institute{School of Computer Science and Engineering,  Sun Yat-sen University, China \and
Huawei Technologies Co., Ltd., China \and
Key Laboratory of Machine Intelligence and Advanced Computing, Ministry of Education, China \and
Guangdong Province Key Laboratory of Information Security Technology, Sun Yat-sen University, Guangzhou\\
\email{gaoyp23@mail2.sysu.edu.cn}, \email{yanglx9@mail.sysu.edu.cn}, \email{\{huangyunmu,xiesong5,lishiyong\}@huawei.com}, \email{wszheng@ieee.org}}
%\end{comment}
%******************

%****************
\newcommand{\jca}{\textcolor[rgb]{1,0,0}}
%****************

\maketitle
\begin{abstract}    
      Under the domain shift, cross-domain few-shot object detection aims to adapt object detectors in the target domain with a few annotated target data. 
      There exists two significant challenges: 
      (1) Highly insufficient target domain data;
      (2) Potential over-adaptation and misleading caused by inappropriately amplified target samples without any restriction.
      To address these challenges, we propose an adaptive method consisting of two parts. First, we propose an adaptive optimization strategy to select augmented data similar to target samples rather than blindly increasing the amount. 
      Specifically, we filter the augmented candidates which significantly deviate from the target feature distribution in the very beginning.
      Second, to further relieve the data limitation, we propose the multi-level domain-aware data augmentation to increase the diversity and rationality  
      of augmented data, which exploits the cross-image foreground-background mixture. 
      Experiments show that the proposed method achieves state-of-the-art performance on multiple benchmarks. The code is available at \href{https://github.com/Hlings/AcroFOD}{https://github.com/Hlings/AcroFOD}.
\footnote{*~indicate the corresponding author.}
\end{abstract}

\section{Introduction}
\label{sec:intro}

Due to the domain discrepancy, apparent performance drop is common when applying a trained detector in an unseen domain.
%Detectors will always suffer a large performance drop while testing on unseen domains because of inevitable domain discrepancy. 
%To address the domain discrepancy problem, it is significant to resort to domain adaptation.
Recently, many researchers try to address it as a domain adaption task.
As one of the domain adaption adaptation sub-tasks, cross-domain few-shot object detection is proposed with the observation that a few samples can still reflect the major characteristics changes of domain shifts\cite{tzeng2017adversarial}, such as view variations\cite{geiger2012kitti,cordts2016cityscapes}, weather diversification \cite{sakaridis2018semantic} and lighting difference\cite{lin2014coco, johnson2017sim10k, amato2019viped}. 
Different from unsupervised domain adaptation (UDA), few labeled target samples are available in few-show domain adaptation (FDA) setting, as well as a large amount samples from source domain.
%the detector won't receive information about new classes but focuses on the robustness of different input distribution.
%Different from few-shot object detection (FSOD), the detector won't receive information about new classes but focuses on the robustness of different input distribution.
% Specifically, cross-domain few-shot object detection is under the few-shot domain adaptation (FDA) setting, where a large amount of data from source domain and only a few labeled data from target domain can be utilized.  

Existing methods\cite{wang2019few, zhuang2020ifan} in the FDA setting are feature-aligning based methods that first pre-train the model in source domain and then align to the target domain.
%need pre-training in source domain and aligning features in the target domain.
Besides, some methods mainly overcome domain gaps under UDA setting\cite{gopalan2011domain, rezaeianaran2021seeking, ramamonjison2021simrod, xu2020exploring}.
%These methods 
However, UDA methods depend on labeled data from the source domain and sufficient unlabeled data from the target domain to fully describe the distribution for both domains, before any explicit or implicit feature alignment operation. %so that they can either explicitly or implicitly align the distribution of the two domains.
%--------------------------------figure introduction--------------------------------------%
\begin{figure}[tbp]
  \centering
  \includegraphics[totalheight=2.0in]{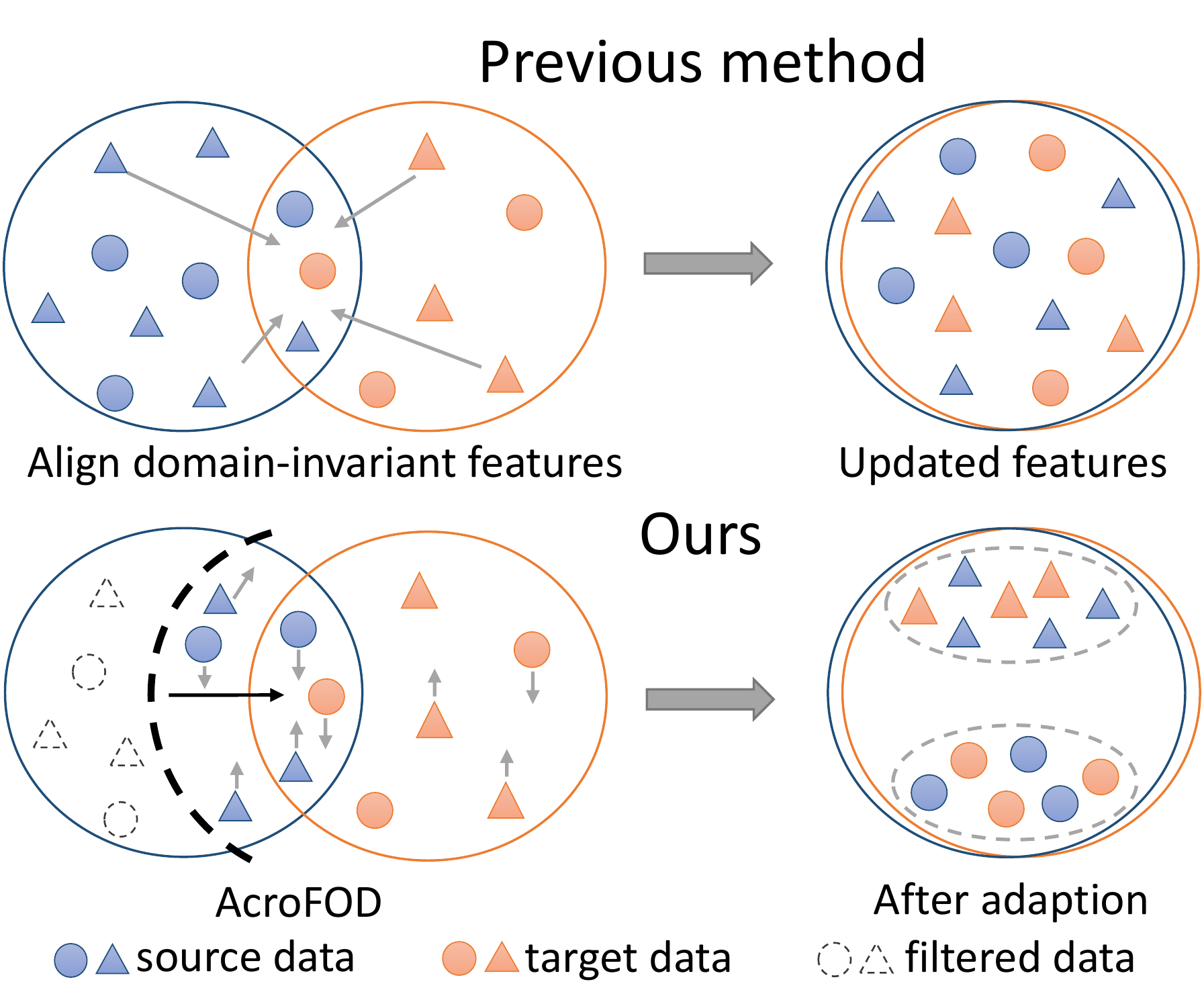}
   \caption{We address the task of cross-domain few-shot object detection. Top: Existing feature-aligning based methods fail to extract discriminative features within limited labeled data in the target domain. Bottom: Our method filters (thick black dotted line) source data that is far away from the target domain. }
   \label{fig:intro}
\end{figure}
%----------------------------------------------------------------------------------------%

However, a large amount of unlabeled data may not available in the target domain\cite{tzeng2017adversarial}. %Meanwhile, this is not workable in FDA if UDA methods are adopted for adaptation because of lacking 
Without sufficient data, most UDA methods performs disappointingly under the FDA setting \cite{chen2018closerclassification, yue2021prototypicalclassification, zou2019consensusdomainadaptation, chen2021dualfewshot, chen2018lstdlowshotobjectdetection, michaelis2018oneshotsegmentation, hu2019attentionlowshotsegmentation, zhang2019fewsegmentation}. An intuitive way to overcome such a problem is to incorporate limited target data with source data and augment them. 
However, we argue that not all the augmented data are useful.
Blind augmentation may even exacerbate the domain discrepancy due the samples that plays as outliers of target data distribution. To overcome this problem, the adaptive optimization of directive augmentation towards target data distribution should be considered very carefully.

%Unfortunately, previous methods almost only focus on augmenting data in single domain, which may cause pre-training on the source domain to over-fit.
%architecture-agnostic and generic paradigm to train a robust detector for cross-domain few-shot object detection. 
In this work, we present an \textbf{A}daptive method for \textbf{Cro}ss-domain \textbf{F}ew-shot \textbf{O}bject \textbf{D}etection (AcroFOD), which is architecture-agnostic and generic. The AcroFOD mainly consists of two parts: an adaptive and iterative distribution optimization for augmented data filtering and multi-level domain-aware augmentation. With a large amount of data available in the source domain, it could be intuitive to train the detector using the whole set of images from source and target domains. However, we argue that such an ungrounded training method is likely to introduce much unsuitable and low-quality data relative to the target domain, which will mislead the model during the training process. To deal with this issue, we design an adaptive distribution optimization strategy to eliminate unsuitable introduced images as shown in Fig. \ref{fig:intro}. Such a strategy allows the detector to fit the feature distribution of target domain faster and more accurately. Moreover, as both background and foreground information can reflect the characteristic of the target domain, we propose multi-level domain-aware augmentation to make a fusion of source and target domain images more diverse and rational.

There are several advantages of the proposed AcroFOD for cross-domain few-shot object detection: 1) \textbf{Wide application scenarios}. In contrast to most of the previous methods\cite{wang2019few, rezaeianaran2021seeking, fan2020fewattention, xu2020exploring, fan2020fewattention}, our method requires neither complicated architecture changes nor generative models for creating additional synthetic data; 2) \textbf{Fast convergence}. The AcroFOD is almost $2\times$ faster than existing methods\cite{ramamonjison2021simrod} to reach better performance in some established scenarios because no pre-training phase is required; 3) \textbf{Less cost of collecting data}. Compared with UDA methods\cite{gopalan2011domain, rezaeianaran2021seeking, ramamonjison2021simrod, xu2020exploring}, we greatly reduce the cost of collecting massive amounts of data in the target domain but introduce the cost of annotation. 

We conduct a comprehensive and fair benchmark to demonstrate the effectiveness of AcroFOD to mitigate different kinds of domain shifts. Our method can achieve new state-of-the-art results on these benchmarks in the FDA setting. The main contributions of this work are summarized as follows:
\begin{itemize}
    \item The proposed adaptive optimization strategy attaches importance to the quality of augmentation. It also prevents the model from over-adaptation which is similar to over-fitting because of lacking data.
    %to tackle the problem of lacking data and ensure the rationality of expansion in cross-domain few-shot object detection task. 

    \item To enhance the diversity of merging images from source and target domain, we construct generalized formulations of multi-level domain-aware augmentation. Then, we provide several instances and discuss them.
\end{itemize}

\section{Related Work}
Existing UDA methods leverage a large number of unlabeled images from the target domain to explicitly mitigate the domain shift. They can be divided into domain-alignment\cite{zhu2019adapting, chen2020harmonizingudaalighment, he2019multiudaalignment, saito2019strongudaalignment}, domain-mapping\cite{ganin2015unsupervised, chen2018domain, kim2019udadiversify} and self-labeling techniques\cite{ramamonjison2021simrod}. 
Zheng et al. \cite{zheng2020udacorsetofine}, propose a hybrid framework to minimize L2 distance between single-class specific prototypes across domains at instance-level and use adversarial training at image-level. ViSGA\cite{rezaeianaran2021seeking} uses a similarity-based grouping scheme to aggregate information into multiple groups in a class agnostic manner. To overcome pseudo-label noise in self-labeling, \cite{ramamonjison2021simrod} proposed a three-step training method with domain-mixed data augmentation, gradual self-adaptation and teacher-guided finetuning.
%relies on the exploration of the law of the data to formulate learning strategies, which 
%Few-shot learning \cite{fei2006oneshotfew} aims to learn new categories with a few data. Meta learning has been proven to achieve satisfactory results in few-shot learning\cite{sun2019metatransfer, zhang2019fewsegmentation, elsken2020metaneural, zhang2021meta, chen2021metabaseline}. Few-shot object detection extends the few-shot learning task from classification to object detection. Recent works have introduced the attention mechanism\cite{fan2020fewattention, chen2021dualfewshot}, multi-relation\cite{wu2020fewmultirelation} and contrastive learning\cite{sun2021fscefewshot} to better capture the characteristics of the new class. 

In the FDA scenario, we expect the model to overcome the domain discrepancy and performance drop due to domain shift in the target domain with only a few target domain data available. In \cite{tzeng2017adversarial}, adversarial learning is used to learn an embedded subspace that simultaneously maximizes the confusion between two domains while semantically aligning their embedding. In cross-domain few-shot object detection, Wang et al. \cite{wang2019few} first adopted a pairing alignment mechanism to overcome the issue of insufficient data. Different from the perspective of modifying the model structure that fails to transfer on other ones, we focus on optimizing the enlarged target data distribution with source data distribution adaptively.
%conduct domain-aware augmentation as a cost-free way to expand data in target domain. 

Data augmentation is an effective technique for improving the performance of deep learning models. Such techniques are mainly divided into two aspects: image-level\cite{yun2019cutmix, zhang2017mixup, yolov4} and box-level\cite{dwibedi2017cut} with pixel-level label\cite{ghiasi2021simple, dvornik2018modeling, fang2019instaboost}. Some other methods consider the combination of multiple geometric and color transformations\cite{hendrycks2019augmix, hendrycks2021deepaug}, while search strategy can find appropriate collocation of them\cite{cubuk2019autoaugmentAA,lim2019fastaa}. Recent works\cite{ramamonjison2021simrod,dacs_domainmix, contrast_domainmix} apply the mixing images technique in cross-domain scenarios. We further propose formulations of both image-level and box-level domain-aware augmentation and conduct them as a cost-free way to generate data between domains diversely.

Our FDA setting follows the prior work\cite{wang2019few}, which prompts model to have stronger generalization ability with only a few samples of target domain.

\section{Approach}
\label{sec:method}
In this section, we present the details of the proposed adaptive method for cross-domain few-shot object detection (AcroFOD). First, we adaptively and iteratively optimize the distribution of candidates towards the target domain for training a robust detector. Then, we generate a lot of candidates to address the problem of insufficiency and sameness of target augmented samples with the proposed multi-level domain-aware augmentation.

The proposed method is motivated by the observation that limited data can still reflect the major characteristics of the target domain\cite{wang2019few}. 
To deal with the lack of data in the target domain, the AcroFOD comprises an adaptive optimization strategy with cross-domain augmentation for reasonable data expansion to overcome domain shifts. 
Sec. \ref{sec:adaptive_optimization} presents our adaptive optimization strategy to promise that augmented target data with source data approximately follow the distribution of target domain. Sec. \ref{sec:da_aug} introduces the formulation of multi-level domain-aware augmentation. Finally, Sec. \ref{sec:acrof_algo} summarizes the whole iterative training process of the AcroFOD.

\begin{figure*}[!htb]
  \centering
  \includegraphics[totalheight=2.4in]{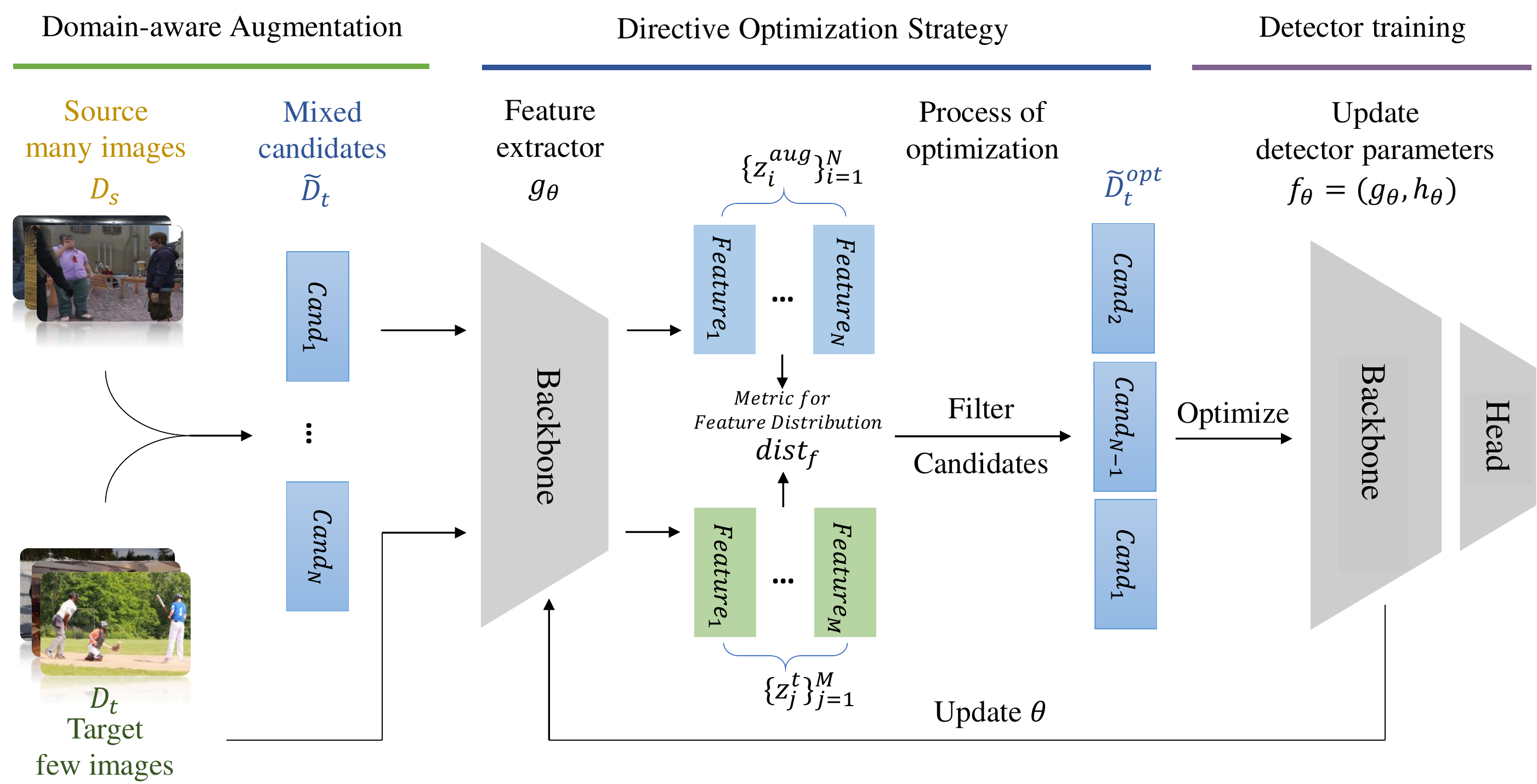}
   \caption{The AcroFOD includes the following three steps. (1) Using the proposed multi-level domain-aware augmentation, we can expand the data distribution of the target domain. (2) Then, the augmented and target domain data is fed into the backbone of the detector to obtain the extracted feature vector. Through the directive optimization strategy, the samples which are unsuitable for mitigating domain shifts can be filtered. (3) Finally, optimized samples will help the detector mitigate domain shifts.}
   \label{fig:DaFOD}
\end{figure*}

\subsection{Problem Statement}
Suppose we have a large data set $D_s=\{(x_i^s, y_i^s)\}_{i=1}^{n_s}$ from the source domain and a few examples $D_t=\{(x_j^t, y_j^t)\}_{j=1}^{n_t}$ from the target domain, where $x_i^s, x_i^t \in \mathcal{X}$ are input images, $y_i^s, y_j^t \in \mathcal{Y}$ consist of bounding box coordinates and object categories for $x_i^s$ and $x_j^t$. We consider scenarios in which there exists the discrepancy between the input source distribution $\mathcal{P}_s:\mathcal{X}\times\mathcal{Y}\rightarrow\mathbb{R}^{+}$ of $D_s$ and target distribution $\mathcal{P}_t:\mathcal{X}\times\mathcal{Y}\rightarrow\mathbb{R}^{+}$ of $D_t$.  

Our goal is to train an adaptive detector $f:\mathcal{X}\rightarrow\mathcal{Y}$ which can alleviate performance drop due to domain gap. However, it is difficult for $f$ to capture domain invariant representation with only a few data $D_t$. 
To effectively exploit the limited information of annotations, we extend $D_t \sim \mathcal{P}_t$ with $D_s \sim \mathcal{P}_s$ to $\widetilde{D}_t$. Supposing $\widetilde{D}_t$ is sampled discretely in assumption distribution $\mathcal{P}_{aug}:\mathcal{X}\times\mathcal{Y}\rightarrow\mathbb{R}^{+}$, we are able to approximate $\mathcal{P}_t(y|x) $ with $\mathcal{P}_{aug}(y|x)$. In fact, we assume $\mathcal{P}_s(y|x) = \mathcal{P}_t(y|x) = \mathcal{P}_{aug}(y|x)$ but $\mathcal{P}_s(x) \ne \mathcal{P}_t(x) \ne \mathcal{P}_{aug}(x)$. It's obvious that some noisy data in $\widetilde{D}_t$ which are dissimilar to $D_t$ may weaken the generalization ability of $f$. 
% in target domain | used for add new lines
In the following subsections, we present the details of AcroFOD.   

\subsection{Adaptive Optimization for Directive Data Augmentation}
\label{sec:adaptive_optimization}
As shown in Fig. \ref{fig:DaFOD}, we are able to generate a bunch of data $\widetilde{D}_t = Aug(D_s, D_t)$ with the introduced domain-aware augmentation which we will discuss later. We expect $\widetilde{D}_t = \{(x_i^{aug}, y_i^{aug})\}_{i=1}^{n_a} \sim \mathcal{P}_{aug}$ to approximate distribution $\mathcal{P}_{t}$ as close as possible. We assume that the detector $f_{\theta}=(g_{\theta},h_{\theta})$ is defined by a set of parameters $\theta$ and consists of backbone $g_{\theta}$ and head $h_{\theta}$. The AcroFOD uses $g_{\theta}$ as feature extractor to output representations of $x_j^t$ in $D_t$, $x_i^{aug}$ in $\widetilde{D}_t$ as follows:

\begin{equation}
\label{eq:feature_extraction}
z_i^{aug} = g_{\theta}(x_i^{aug}), z_j^t = g_{\theta}(x_j^t).
\end{equation}

Then, the AcroFOD sorts augmented candidates $x_i^{aug}$ according to the distance of representations between $x_i^{aug}$ and $\{x_j^t\}_{j=1}^{n_t}$ measured by metric function $dist_f$:

\begin{equation}
\label{eq:distance}
d_i^{aug} = dist_f(z_i^{aug},\{z_j^t\}_{j=1}^{n_t}).
\end{equation}

In order to filter a certain amount of noisy samples in $\widetilde{D}_t$, we use shrinkage ratio $k$ ( $0<k\le 1$ ) to decrease the quantity of expanded candidates. Then, we define an optimization function $\phi_{opt}$ to optimize $\widetilde{D}_t$ with $d_i^{aug}$, resulting in the optimized extended domain $\widetilde{D}_t^{opt}$ defined as follows:

\begin{equation}
\label{eq:opt_function}
\widetilde{D}_t^{opt} = \{(x_i^{opt}, y_i^{opt})\}_{i=1}^{n_b} = \phi_{opt}(\widetilde{D}_t, \{d_i^{aug}\}_{i=1}^{n_a}, k).
\end{equation}

Through $\phi_{opt}$, top $n_b$ ($n_b = \lfloor n_a \ast k \rfloor$) candidates are chosen from $\widetilde{D}_t$ in increasing order of $d_i^{aug}$. With $\phi_{opt}$, we can obtain $\widetilde{D}_t^{opt}$ to better reflect the target domain distribution $\mathcal{P}_{t}$. However, such suitable $\widetilde{D}_t^{opt}$ is likely to change as $f_{\theta}$ converges. To tackle this problem, we optimize $\widetilde{D}_t^{opt}$ iteratively. Given detectors $f_{\theta}^{a} $ and $f_{\theta}^{b}$ trained after $a$ and $b$ epochs $(a > b \ge 0)$ during training process. The error of $f_{\theta}$ in source and target domain $\epsilon_{D_s}(f_{\theta}^a), \epsilon_{D_t}(f_{\theta}^a)$ are expected to be smaller than $\epsilon_{D_s}(f_{\theta}^b), \epsilon_{D_t}(f_{\theta}^b)$ due to $g_{\theta},h_{\theta}$ updating. So, $g_{\theta}^{a}$ is able to represent $x_i^{aug}$ and $x_i^t$ more accurately than $g_{\theta}^{b}$. 
Then, we can iteratively optimize $\widetilde{D}_{t}^{opt}$ by $dist_f$ with updating feature representation $z_i^{aug}$ and $z_i^t$.

At the $n$th ($n \ge 1$) epoch, $\widetilde{D}_{t^n}^{opt}$ can be obtained by filtering $\widetilde{D}_t$ as follows:

\begin{equation}
\label{eq:iterative_updating}
\widetilde{D}_{t^n}^{opt} = \phi_{opt}(\widetilde{D}_t, \{dist_f(g_{\theta}^{n}(x_i^{aug}),\{g_{\theta}^n(x_j^t)\}_{j=1}^{n_t})\}_{i=1}^{n_a}, k).
\end{equation}

Finally, the adaptive detector $f_{\theta}^{n}=(g_{\theta}^n,h_{\theta}^n)$ can also be optimized iteratively by $(x_i^{opt}, y_i^{opt}) \in \widetilde{D}_{t^n}^{opt}$ as follows:

\begin{equation}
\label{eq:opt_process}
g_{\theta}^{n+1},h_{\theta}^{n+1} \leftarrow optimizer((g_{\theta}^n,h_{\theta}^n), \nabla_{\theta}\mathcal{L}_{\theta}(f_{\theta}^n(x_i^{opt}),y_i^{opt}), \eta),
\end{equation}

\noindent where the $optimizer$ is an optimizer, $\eta$ is the learning rate for $g_{\theta}, h_{\theta}$ and $\mathcal{L}$ is the loss function. 

We intend to measure the correlation between $z_i^{aug}$ and $\{z_j^t\}_{j=1}^{n_t}$. Therefore, we utilize two widely-used metric functions $MMD, CS$ as $dist_f$ .

\noindent\textbf{First: Maximum Mean Discrepancy.} The Maximum Mean Discrepancy (MMD)\cite{gretton2012kernelmmd} distance is used to measure the distance of these two distributions in the Reproducing Keral Hilbert Space (RKHS). For $z_i^{aug}$, $\{z_i^t\}_{i=1}^{n_t}$ defined in Eq. \ref{eq:feature_extraction}, $dist_f$ is instantiated to $MMD^2$ as:
\begin{equation}
\label{eq:mmd_dis}
MMD^2(z_i^{aug}, \{z_j^t\}_{j=1}^{n_t}) = ||\frac{1}{n_t} \sum_{j=1}^{n_t} z_j^t - z_i^{aug}||_2^2.
\end{equation}

\noindent\textbf{Second: Cosine Distance.} Cosine distance is an effective metric to measure the similarity of samples in the embedding space \cite{luo2019cosine1, guo2020multicosine, chen2019progressivecosine}. Eq. \ref{eq:distance} can be rewritten as $CS$ in the following expression:
\begin{equation}
\label{eq:cosine_dis}
CS(z_i^{aug}, \{z_j^t\}_{j=1}^{n_t}) = \sum_{j=1}^{n_t}(1 - \frac{z_j^t\cdot z_i^{aug}}{||z_j^t||_2\cdot||z_i^{aug}||_2}).  
\end{equation}

\subsection{Multi-level Domain-aware Augmentation}
\label{sec:da_aug}
From Sec. \ref{sec:adaptive_optimization}, the convergence of $f_{\theta}$ relies on the ${D}_{t}^{opt}$ optimized by $D_t^{aug}=Aug(D_s,D_t)$. The simple combination of $D_s$ and $D_t$ limits the variety of $D_t^{aug}$. To generate more adequate samples while controlling the overhead of training computation, we propose domain-aware augmentation as $Aug$ at \textbf{image-level} and \textbf{box-level}.
Here, we give uniform formulations for each level of $Aug$ and then provide several specific instantiations of them.

\noindent\textbf{Image-level Domain-aware Augmentation.} Given a batch of data in source domain  $B_s=\{(x_i^s,y_i^s)\}_i^{n_{bs}}$ and target domain $B_t=\{(x_i^t,y_i^t)\}_i^{n_{bt}}$. We sample $m \le n_{bs}$ and $n \le n_{bt}$ data from $B_s$ and $B_t$ from these two domains respectively. Then, we randomly mix them to a single image $x^{aug}$ as follows:
\begin{equation}
\label{eq:x_aug_imagelevel}
\begin{aligned}
&x^{aug} = x^{aug}_0 + \sum_{i=1}^{m}\sum_{j=1}^{n}A_{(i,j)}(\lambda_ix_i^s + \lambda_jx_j^t),
\end{aligned}
\end{equation}

\noindent where $x^{aug}_0$ is an initialized empty image whose size is different from both $x_i^s$ and $x_i^t$, $A_{(i,j)}$ is the hand-crafted transformation matrix for image pair$\{x_i^s,x_j^t\}$. $\lambda_i$ and $\lambda_j$ $(1 \ge \lambda_i + \lambda_j \ge 0)$ are the corresponding weights of $x_i^s$ and $x_j^t$, respectively. Then, we can recompute the label set $y^{aug}=\{(y_{box}^{aug},y_{cls}^{aug})\}$ of $x^{aug}$ as follows:
\begin{equation}
\label{eq:y_aug_imagelevel}
\begin{aligned}
y_{box}^{aug} &= \underset{i=1...m,j=1...n}{Concat}(A_{(i,j)}^Ty_{i(box)}^{s}, A_{(i,j)}^Ty_{j(box)}^{t}),\\
y_{cls}^{aug} &= \underset{i=1...m,j=1...n}{Concat}(\lambda_i^{cls}y_{i(cls)}^{s} + \lambda_j^{cls}y_{j(cls)}^{t}),
\end{aligned}
\end{equation}
where $y_{i(box)}^{s}$ and $y_{j(box)}^{t}$ denote the bounding box coordinates of interest instances from source and target domains. $y_{i(cls)}^{s}$ and $ y_{j(cls)}^{t}$ represent corresponding confidences of categories. $\lambda_i^{cls}$ and $\lambda_j^{cls}$ are weights of the corresponding confidence scores.

With Eq. \ref{eq:x_aug_imagelevel} and Eq. \ref{eq:y_aug_imagelevel}, we describe two versions of image-level domain-aware augmentation. First, we define $m+n=4 (m,n\in \mathbb{N}^{+})$, $\lambda_i=\lambda_i^{cls}$ and $\lambda_i | \lambda_j = 1$ as \underline{\textit{domain-splice}}. Second, to increase the degree of interaction at the image-level, we choose $m+n=2 (m,n\in \mathbb{N}^{+})$ and then weight two images with $\lambda_i + \lambda_j = 1, 1 \ge \lambda_i,\lambda_j \ge 0$, $\lambda \sim Beta(\alpha, \alpha)$ and $\lambda_i \& \lambda_j = 1$ as \underline{\textit{domain-reallocation}}. The above two methods can also be combined to generate more diverse images. 

\noindent\textbf{Box-level Domain-aware Augmentation.} To effectively utilize limited instance annotation, we can separate them from the background and then put them on the other regions. In order to improve the generality of proposed augmentations, we focus on utilizing domain-aware box-level labels rather than pixel-level labels which are often used in previous works\cite{ghiasi2021simple, dvornik2018modeling, fang2019instaboost}. Here, we propose the formulation of box-level domain-aware augmentation with bounding box labels.

For bounding box $b^s$ and $b^t$ from source and target domain with the resized width $w$ and height $h$, we exchange them to combine the characteristic of each other. The formulation is presented as follows:
\begin{equation}
\label{eq:x_aug_box_level}
\begin{aligned}
b_{(p,q)}^{aug} &= \beta_{(p,q)}b_{(p,q)}^s + (1-\beta_{(p,q)})b_{(p,q)}^t,
\end{aligned}
\end{equation}
where $(p,q),p=1,2...w,q=1,2...h$ represents the index of pixels in box $b^s$ and $b^t$. $\beta(p,q) \in [0,1]$ is the corresponding weight for each index. 
From Eq. \ref{eq:x_aug_box_level}, we denote $\beta(p,q) = 0(\forall p,q)$ as \underline{\textit{direct}} exchange and define $\beta(p,q) =\beta_{mix}, \beta_{mix} \sim Beta(\alpha_m,\alpha_m)$ as \underline{\textit{mixture}} exchange, where $\alpha_m$ is a hyper-parameter.

\begin{figure}[tb]
  \centering
  \includegraphics[totalheight=1.4in]{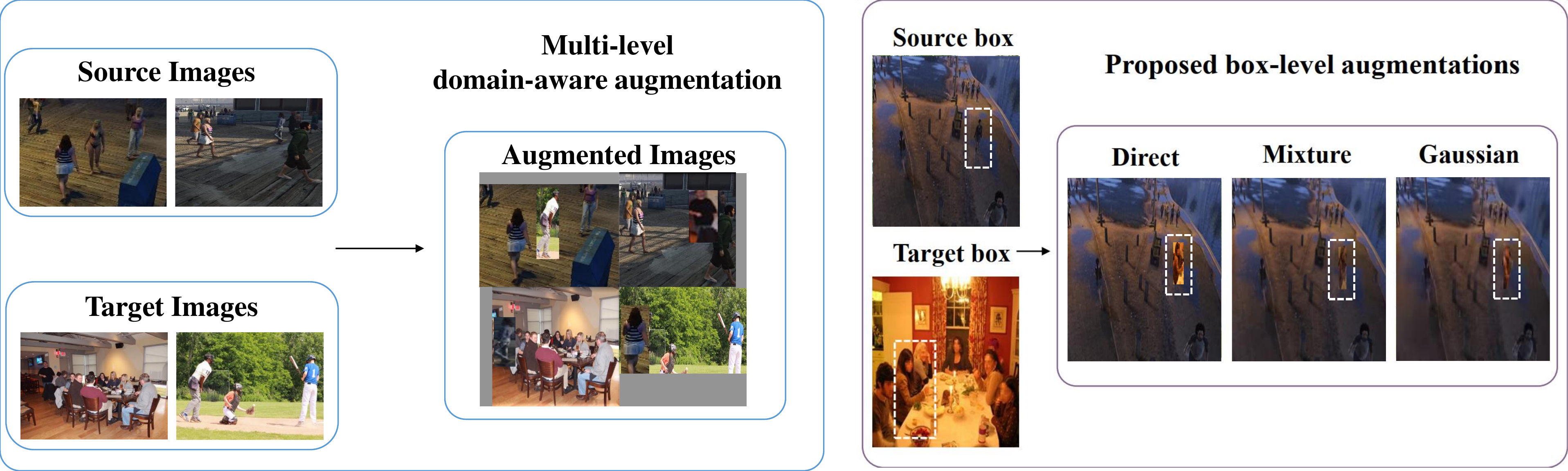}
   \caption{Left: visualization of multi-level domain-aware augmentation. Right: three types of proposed box-level cross-domain augmentations.}
   % refine caption
   \label{fig:gaussian_cp}
\end{figure}

Instances under different scales have different degrees of dependence on context information\cite{chen2021scaleaware}. In order to obtain scale-aware weights for each pixels, we propose \underline{\textit{Gaussian}} exchange which adopts the Gaussian map defined as follows:

\begin{equation}
\begin{aligned}
&\beta(p,q) = exp(-(\frac{(p-\mu_x)^2}{\sigma_x^2} + \frac{(q-\mu_y)^2}{\sigma_y^2})), \\
&\sigma_x = \frac{w}{W}\sqrt{\frac{hw}{2\pi}} \quad \sigma_y = \frac{h}{H}\sqrt{\frac{hw}{2\pi}},
\end{aligned}
\end{equation}

\noindent where $H$, $W$ are height and weight of the image, $\sigma_x, \sigma_y$ are the variance of $x$ and $y$ axes in box $b^s$ and $b^t$. $\mu_x,\mu_y$ are the corresponding mean in box $b^s$ and $b^t$. 
The comparison of three box-level augmentations is shown in Fig. \ref{fig:gaussian_cp}.

\subsection{Cross-domain Training Framework}
\label{sec:acrof_algo}
We expect that the augmented samples are still closer to the distribution of the target domain. 
Therefore, these data will be fed into the backbone network of the target detector to calculate features, and then suitable data will be selected according to the distance from the target domain. The feature extractor used for obtaining $\widetilde{D}_t^{opt}$ will update after every epoch. 
Note that our model, including the backbone and the detector module, is trained from scratch. Details of our AcroFOD are presented in \textbf{Algo.} \ref{CDTF}.

\begin{algorithm}
\caption{Adaptive Method AcroFOD}
\label{CDTF}
  \begin{algorithmic}
    \Require
      Initialized detector $ \theta^{ini} $,
      the source domain $D_s=\{(x_i^s, y_i^s)\}_{i=1}^{n_s}$,
      few data target domain $D_t=\{(x_j^t, y_j^t)\}_{j=1}^{n_t}$,
      total epochs $T$, distance function $dist_f$,
      amount of steps for every epoch $N$,
      domain-aware augmentation function $Aug$,
      shrinkage ratio $k$,
      loss function $\mathcal{L}$.
    \Ensure Adaptive Detector $ f_{\theta} = (g_{\theta}, h_{\theta})$
    \State Initialize $\theta \leftarrow \theta^{ini}$
    \State Initialize feature extractor $g_{\theta}$
    \For {$epoch \leftarrow 1,...,T $}
        \State $\widetilde{D}_t = \{(x_i^{aug},y_i^{aug})\}_{i=1}^{n_a} = Aug(D_s,D_t)$
        \State $\widetilde{D}_{t}^{opt} = \phi_{opt}(\widetilde{D}_t, \{dist_f(g_{\theta}(x_i^{aug}),\{g_{\theta}(x_j^t)\}_{j=1}^{n_t})\}_{i=1}^{n_a}, k)$
        \For {$step \leftarrow 1,...,N $}
            \State sample batch $B=\{(x_i^{opt},y_i^{opt})\}_{i=1}^{bs}$ from $\widetilde{D}_t^{opt}$ 
            \State $pred = f_{\theta}(\{x_1^{opt},...,x_{bs}^{opt}\})$
            \State $loss = \mathcal{L}(pred, \{y_1^{opt},...,y_{bs}^{opt}\})$
            \State Update $\theta$ to minimize $loss$
        \EndFor
        \State $g_{\theta},h_{\theta} \leftarrow \theta$
    \EndFor
  \State $ f_{\theta} \leftarrow \theta$
  \end{algorithmic}
\end{algorithm}
%-------------------------------------------------------------------------
% section four
\section{Experiments}
We present the results of proposed adaptive method for cross-domain few-shot object detection (AcroFOD) in various scenarios like adverse weather, synthetic-to-real and cross-camera domain shifts in Sec. \ref{sec:all_benchmark}. Then, we provide qualitative results of adverse weather benchmark in Sec. \ref{sec:qual_result}. Furthermore, we analyze effect of multiple parts of the AcroFOD in Sec. \ref{sec:abl_study}. Finally, we explore the performance of our method on different data magnitudes in Sec. \ref{sec:more_data}.
\subsection{Experimental Setup}
\label{sec:exp_introduce}
\noindent\textbf{- Adverse Weather Benchmark (C $\rightarrow$ F)}. In this scenario, we use Cityscapes\cite{cordts2016cityscapes} as the source dataset. It contains 3,475 real urban images, with 2,975 images used for training and 500 for validating. Foggy version of Cityscapes\cite{sakaridis2018semantic} is used as the target dataset. Highest fog intensity (least visibility) images of 8 different categories are used in our experiments, matching prior work\cite{xu2020cross}. Following \cite{wang2019few}, we used the tightest bounding box of an instance segmentation mask as ground truth box. This scenario is referred to C $\rightarrow$ F.

\noindent \textbf{- Synthetic-to-real Benchmark (S $\rightarrow$ C, V $\rightarrow$ O).} SIM10k\cite{johnson2017sim10k} is a simulated dataset that contains 10k synthetic images. In this dataset, we use all 58,701 car bounding boxes available as the source data during training. For the target data and evaluation, we use Cityscapes\cite{cordts2016cityscapes} and only consider the car instances. This scenario is referred to S $\rightarrow$ C. ViPeD\cite{amato2019viped} contains 200K frames collected from the video game with bounding box annotations for person class. We select one frame per 10 frames and a total of 20K frames as the source dataset. We select COCO\cite{lin2014coco} as the target dataset and only consider the person class. We denote this scenario as V $\rightarrow$ O.

\noindent\textbf{- Cross-camera Benchmark (K $\rightarrow$ C).} In this scenario, we use the KITTI\cite{geiger2012kitti} as our source data. KITTI contains 7,481 images and we use all of them for training. Similar to the previous scenarios, we use Cityscapes\cite{cordts2016cityscapes} as the target data. Following prior works\cite{rezaeianaran2021seeking, ramamonjison2021simrod}, only the car class is used. This scenario is abbreviated as K $\rightarrow$ C.
% We compare the AcroFoD with other popular methods. 

\noindent\textbf{- Implementation Details.} We adopt the single-stage detector YOLOv5\cite{yolov5} as the baseline and compare with unsupervised domain adaptation (UDA) and few-shot domain adaptation (FDA) methods at the same time. For UDA setting\cite{rezaeianaran2021seeking}, we report their results based on the full amount of target domain data. For FDA setting\cite{wang2019few}, we report mean and deviation for 5 rounds using the same number of images. Meanwhile, we also compare our method with proportional sampling (denote as "proportion") which samples data from source and target domains uniformly for training and few-shot object detection method FsDet\cite{wang2020fsdet} in the FDA setting. In all experiments, we adopt adaptive optimization strategy from scratch and set shrinkage ratio $k=0.8$. The effect of $k$ will analyze in Sec. \ref{sec:abl_study}. For a fair comparison, we resize input images to $640 \times 640$ in all experiments without any extra dataset (such as COCO\cite{lin2014coco}) for model pre-training.
For evaluation metrics, We denote average precision with IoU threshold of 0.5 as AP50 for a single class or mAP50 for multi classes, and AP or mAP for 10 averaged IoU thresholds of 0.5:0.05:0.95\cite{lin2014coco}.

\subsection{Main Results} 
%%%%%%%%%%%%%%%%%%%%%%%%%%%%%%%%%%%    Cityscpaes to foggy exp %%%%%%%%%%%%%%%%%%%%%%%%%%%
\begin{table}[t]
  \caption{Results in C $\rightarrow$ F scenario. "V" and "R" stand for VGG16 and ResNet50 backbone respectively. "X" stands for a type of yolov5 model. In FDA setting, only 8 fully annotated images are used for domain adaption per round. $^*$ and $^\dag$ represent only using optimization and augmentation respectively. The last row combines both of them.}
  \centering
  \resizebox{\textwidth}{23mm}{
  \begin{tabular}{c|cc|cccccccc|cc}
    \hline
    Setting & Method    & Arch. & person & rider & car & truck & bus & train & mcycle & bicycle & mAP50 & gain \\
    \hline
            &source     & V &  24.1  &  29.9 & 32.7 &  10.9 &13.8 &  5.0  &  14.6  & 27.9    & 19.9&  -   \\
            &source     & R &  27.2  &  31.8 & 32.5 &  16.0 &25.5 &  5.6  &  19.9  & 27.0    & 22.8&  -   \\
            &source     & X &  30.8  &  27.8 & 43.7 &  8.2 &24.3 &  4.8  &  11.0   & 24.6    & 21.9&  -   \\
            &Pre+FT & X &  31.2\tiny$\pm0.3$  & 28.1\tiny$\pm0.2$ & 44.1\tiny$\pm0.5$ &  8.3\tiny$\pm0.2$ &24.3\tiny$\pm0.1$  & 5.9\tiny$\pm0.6$  & 11.1\tiny$\pm0.3$ & 24.6\tiny$\pm0.2$ &22.2\tiny$\pm0.2$&  0.3 \\
            &proportion & X &  30.0\tiny$\pm0.4$  & 28.7\tiny$\pm0.7$ & 41.8\tiny$\pm1.2$ &  13.1\tiny$\pm0.5$ &22.3\tiny$\pm0.7$ &  9.6\tiny$\pm1.5$  & 19.3\tiny$\pm1.8$ & 24.7\tiny$\pm0.7$ &23.7\tiny$\pm0.5$&  1.8 \\  
    \hline
    \multirow{4}*{UDA}
            &DA-Faster\cite{chen2018domain}  & V &  25.0 & 31.0 & 40.5 & 22.1  & 35.3 &  20.2 & 20.0  &  27.1  & 27.6 &  7.7   \\
            &FAFRCNN\cite{wang2019few}      &  V &  29.1 & 39.7 & 42.9 & 20.8  & 37.4 &  24.1 & 26.5  &  29.9 & 31.3 &  11.4   \\   
            &SWDA\cite{saito2019strong}      &  R &  31.8 & 44.3 & 48.9 & 21.0  & 43.8&   28.0 & 28.9  &  35.8   & 35.3 &  12.5   \\
            &ViSGA\cite{rezaeianaran2021seeking}&  R & 38.8  & 45.9  & 57.2 & 29.9  & 50.2 &  51.9 &  31.9  & 40.9&\textbf{43.3}& \textbf{20.5}   \\
    \hline
    \multirow{8}*{FDA}
            & ADDA\cite{tzeng2017adversarial}  &  V & 24.4\tiny$\pm0.3$ &29.1\tiny$\pm0.9$& 33.7\tiny$\pm0.5$& 11.9\tiny$\pm0.5$ & 13.3\tiny$\pm0.8$ & 7.0\tiny$\pm1.5$ & 13.6\tiny$\pm0.6$& 27.6\tiny$\pm0.2$ & 20.1\tiny$\pm0.8$ &  0.2 \\
            & D$T_{f}$+FT\cite{johnson2016perceptual}&  V & 23.5\tiny$\pm0.5$   & 28.5\tiny $\pm0.6$ & 30.1\tiny $\pm0.8$&  11.4\tiny $\pm0.6$ & 26.1\tiny $\pm0.9$&   9.6\tiny $\pm2.1$ &  17.7\tiny $\pm1.0$  & 26.2\tiny $\pm0.6$    & 21.7\tiny $\pm0.6$&  1.8 \\
            &DA-Faster\cite{chen2018domain}   &  V & 24.0\tiny $\pm0.8$ & 28.8\tiny $\pm0.7$ & 27.1\tiny $\pm0.7$ & 10.3\tiny $\pm0.7$ & 24.3\tiny $\pm0.8$ & 9.6\tiny $\pm2.8$ &14.3\tiny $\pm0.8$  & 26.3\tiny $\pm0.8$ & 20.6\tiny $\pm0.8$& 0.7   \\
            & SimRoD \cite{ramamonjison2021simrod}  &  X & 34.3\tiny $\pm1.3$ & 35.8\tiny $\pm0.3$  & 55.9\tiny $\pm0.8$& 9.6\tiny $\pm1.8$  & 18.0\tiny $\pm0.6$ & 5.9 \tiny$\pm0.3$ & 10.6 \tiny$\pm0.2$ & 29.2\tiny $\pm0.8$ & 24.9\tiny $\pm0.2$ & 4.7\\
            & FsDet \cite{wang2020fsdet}     &  X & 32.3\tiny $\pm1.2$ & 29.8\tiny $\pm1.2$  & 44.0\tiny $\pm1.7$ & 14.1\tiny $\pm2.2$  & 24.2\tiny $\pm1.4$ & 8.4\tiny $\pm1.2$& 22.9\tiny $\pm1.6$  & 26.2\tiny $\pm2.2$ & 25.2\tiny $\pm1.1$ & 3.3  \\ 
            \cline{2-13}
            & AcroFOD$^*$ & X & 31.8\tiny$\pm0.9$ &30.9\tiny$\pm1.2$  & 43.9\tiny$\pm2.3$ & 15.3\tiny$\pm2.1$ & 27.8\tiny$\pm1.8$  &8.8\tiny$\pm1.3$& 26.2\tiny$\pm1.9$  & 26.3\tiny$\pm0.8$ & 26.4\tiny$\pm1.0$ & 4.5 \\
            & AcroFOD$^\dag$ & X & 36.5\tiny$\pm1.4$ &37.4\tiny$\pm1.3$ & 51.6\tiny$\pm0.9$ & 17.9\tiny$\pm1.1$ & 33.0\tiny$\pm0.7$  &26.4\tiny$\pm1.2$& 27.5\tiny$\pm1.1$  &  31.5\tiny$\pm1.5$ & 32.7\tiny$\pm0.6$ & 10.8 \\
            & AcroFOD & X & \textbf{46.2}\tiny$\pm0.5$ &\textbf{47.3}\tiny$\pm0.6$  & \textbf{63.5}\tiny$\pm0.4$ & \textbf{20.1}\tiny$\pm1.6$ & \textbf{41.5}\tiny$\pm0.8$  &\textbf{34.2}\tiny$\pm1.8$& \textbf{36.1}\tiny$\pm0.7$  &  \textbf{39.6}\tiny$\pm0.9$  & \textbf{41.1}\tiny$\pm0.8$& \textbf{19.2} \\
    \hline
  \end{tabular} }
  \label{tab:city_to_foggy}
\end{table}

\label{sec:all_benchmark}
In this section, we evaluate the proposed method by conducting extensive experiments on the established scenarios. 

\noindent\textbf{- Results for Scenarios C $\rightarrow$ F.} As summarized in Table \ref{tab:city_to_foggy}, our proposed AcroFOD performs significantly better than other compared FDA methods in all categories. Besides, the AcroFOD achieves mAP50 at 41.1\%, which is 19.2\% higher than the baseline method, solely trained on source data.

It is observable that other baseline methods only obtain less improvement for both mAP50 and gain. The compared SimRoD\cite{ramamonjison2021simrod} also employees domain-mix augmentation and improves the generation ability of the used detector. Our AcroFOD also achieves significantly better performance than SimRoD, which indicates that a simple augmentation is not sufficient to train a robust object detector to mitigate the domain gap.  
%---------------------------    Sim10k to Cityscapes & KITTI to Cityscapes exps  -----------------------------%
\begin{table}[t]
  \caption{Results of AP50 for the S $\rightarrow$ C and K $\rightarrow$ C adaptation scenarios. In FDA setting, we randomly choice 8 images in target domain. "Source" refers to the model trained using source data only. "Adaptation" means the model adapted by target data. $^*$ and $^\dag$ represent only using optimization and augmentation respectively. The last row combines both of them.} % fully annotated
  \centering
  \subtable[S $\rightarrow$ C]{
  \scalebox{0.8}{
  \begin{tabular}{c|ccccc}
    \toprule
    Setting & Method & Source    & Adaptation & gain \\
    \midrule
    \multirow{4}*{UDA}
    & FAFRCNN\cite{wang2019few} & 33.5 & 41.2 & 7.7  \\
    & DA-Faster\cite{chen2018domain} & 31.9 & 41.9 & 10.0  \\    
    & SWDA\cite{saito2019strong}  & 31.9 & 44.6 & 12.7 \\
    %& SimRoD\cite{ramamonjison2021simrod} & 55.9 & 62.2 & 6.3  \\     
    & ViSGA\cite{rezaeianaran2021seeking} & 31.9  & 49.3 & \textbf{17.4} \\
    \hline
    \multirow{6}*{FDA} 
    %& ADDA\cite{tzeng2017adversarial}  & 33.5 & 34.4\tiny$\pm0.7$ & 0.9 \\
    %& D$T_{f}$+FT\cite{johnson2016perceptual} & 33.5  & 35.6\tiny$\pm0.6$ & 2.1 \\
    %& FAFRCNN\cite{wang2019few} & 33.5 & 40.3\tiny$\pm0.6$ & 6.8  \\
    & Pre+FT &49.0 & 49.4\tiny$\pm0.3$ &0.4      \\
    & proportion & 49.0 &  50.2\tiny$\pm0.6$ & 1.2 \\
    & FsDet\cite{wang2020fsdet} & 49.0 & 52.9\tiny$\pm1.2$ & 3.9\\
    & SimRoD\cite{ramamonjison2021simrod} & 49.0 & 54.2\tiny$\pm0.5$ & 5.2   \\
    \cline{2-5}
    & AcroFOD$^*$ & 49.0 & 55.6\tiny$\pm2.6$ & 6.6 \\
    & AcroFOD$^\dag$ & 49.0 & 57.4\tiny$\pm2.1$ & 8.4 \\
    & AcroFOD & 49.0 & \textbf{62.5}\tiny$\pm1.6$ & \textbf{13.5} \\
    \bottomrule
  \end{tabular}} 
  \label{tab:sim10k_to_city}}
  \subtable[ K $\rightarrow$ C]{
  \scalebox{0.8}{
  \begin{tabular}{c|cccc}
    \toprule
    Setting & Method & Source     & Adaptation & gain \\
    \midrule
    \multirow{4}*{UDA}
    & DA-Faster\cite{chen2018domain} & 32.5 & 41.8 & 9.3  \\    
    & SWDA\cite{saito2019strong}  & 32.5 & 43.2 & 10.7 \\
    & ViSGA\cite{rezaeianaran2021seeking} & 32.5 & 47.6 & 15.1 \\
    & GPA\cite{xu2020cross} & 32.5 & 47.9 & \textbf{15.3}  \\ 
    \hline
    \multirow{6}*{FDA} 
    & Pre+FT &47.4 & 47.7\tiny$\pm0.3$ &0.3      \\
    & proportion &47.4 & 47.6\tiny$\pm0.2$ &0.2      \\
    & FsDet\cite{wang2020fsdet} &47.4 & 52.9\tiny$\pm1.2$ &5.5    \\
    & SimRoD\cite{ramamonjison2021simrod} & 47.4  & 55.8\tiny$\pm0.6$  & 8.4   \\
    \cline{2-5}
    & AcrFOD$^*$    & 47.4 & 51.9\tiny$\pm2.9$ & 4.5  \\
    & AcrFOD$^\dag$ & 47.4 & 53.9\tiny$\pm1.6$ & 6.5  \\
    & AcrFOD        & 47.4 & 62.6\tiny$\pm2.1$ & \textbf{15.2}  \\
    \bottomrule
  \end{tabular}} 
  \label{tab:kitti_to_city}}
  \label{tab:main_results}
\end{table}
%------------------------------------------------------------------------------------------%
%---------------------------    Viped to 0.1%COCO exp  -----------------------------%
\begin{table}[t]
  \caption{Results of V $\rightarrow$ O adaptation scenario. We randomly select 60 fully annotated images from coco person for each round. $^*$ and $^\dag$ represent only using optimization and augmentation respectively. The last row combines both of them. }
  \centering
  \renewcommand\tabcolsep{2.5pt}  % 调整表格列间的长度
  \begin{tabular}{c|cccc}
    \toprule
     Method & AP50  & AP & gain-AP50 &gain-AP \\
    \midrule
    Source     & 30.4 & 13.0 & -    & - \\
    proportion & 31.8\tiny$\pm1.5$ & 13.5\tiny$\pm0.3$ & 1.4  & 0.5\\
    Pre+FT     & 43.2\tiny$\pm0.8$ & 21.0\tiny$\pm0.5$ & 13.2 & 8.0\\
    \hline
    FsDet\cite{wang2020fsdet} & 36.7\tiny$\pm1.9$ & 15.9\tiny$\pm0.8$ &6.3 &2.9   \\
    SimRoD\cite{ramamonjison2021simrod} & 42.8\tiny$\pm1.0$ & 19.5\tiny$\pm0.7$  & 12.4 & 6.5   \\
    \cline{1-5}
    AcroFOD$^*$    & 42.0\tiny$\pm1.2$ & 19.2\tiny$\pm1.3$ & 11.6 & 6.2  \\
    AcorFOD$^\dag$ & 41.4\tiny$\pm0.7$ & 18.7\tiny$\pm0.6$ & 11.0 & 5.7 \\
    AcroFOD        & \textbf{45.8}\tiny$\pm0.6$ & \textbf{22.5}\tiny$\pm0.4$ & \textbf{15.4} & \textbf{9.5} \\ 
    \bottomrule
  \end{tabular}
  \label{tab:viped_to_0.1coco}
\end{table}
%------------------------------------------------------------------------------------------%

\noindent\textbf{- Results for Other Three Scenarios.} As presented in Table \ref{tab:sim10k_to_city} and Table \ref{tab:kitti_to_city}, results show similar trends with previous evaluation C $\rightarrow$ F. In FDA setting, our AcroFOD performs better than previous methods in single class domain adaptation, such as car and person. Meanwhile, we obtain comparable performance to many UDA methods. 
Other methods in UDA setting sometimes perform better than AcroFOD which only uses 8 target images in FDA setting. 
As shown in Table \ref{tab:viped_to_0.1coco}, our AcroFOD outperforms pre-training + fine-tuning paradigm (denoted as Pre+FT) about 1.7\%AP50 and 0.3\%AP in V $\rightarrow$ O scenario, which suggests our framework can still handle complex domain shift in person class. 

%In summary, the good performance shown by our model across three datasets prove that our framework can well align domain discrepancy with only few target data.

\subsection{Qualitative Results}
\label{sec:qual_result}

\begin{figure}[t] %!htbp
  \centering
  \includegraphics[totalheight=1.6in]{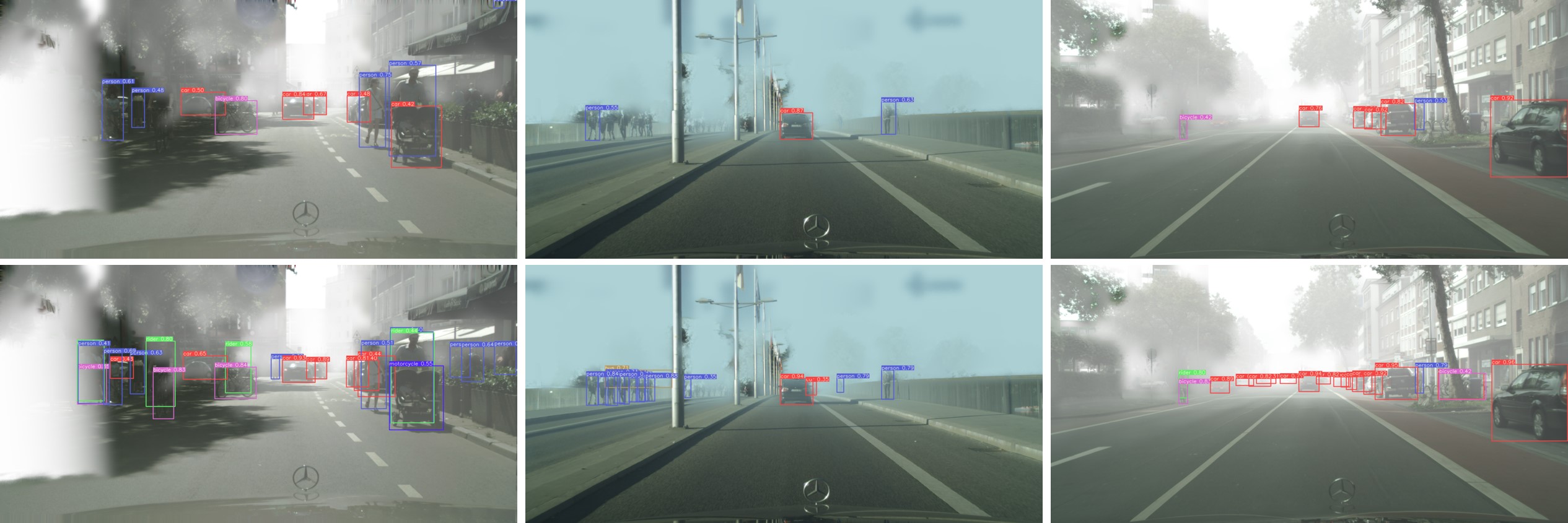}
   \caption{Qualitative result. The results are sampled from C $\rightarrow$ F scenario, we set the bounding box visualization threshold of 0.3. The first/second rows are output results from unadapted/adapted training models respectively.}
   \label{fig:foggy_quat_result}
\end{figure}

Fig. \ref{fig:foggy_quat_result} shows some qualitative results of C $\rightarrow$ F. It can be clearly observed that 
%1) the adapted model outputs tighter bounding boxes for each object, indicating better localization ability; 
1) the AcroFOD motivates the detector to place higher confidence on detected objects, especially for occluded objects; 2) the model after adaptation detects more targets than the one trained with only source data.

\subsection{Ablation Study}
\label{sec:abl_study}
To evaluate the impact of various components of the AcroFOD on detection performance, we use all four scenarios for evaluation. Following prior settings\cite{wang2019few}, in C $\rightarrow$ F, S $\rightarrow$ C and K $\rightarrow$ C adaptation scenarios, we randomly sample 8 images in the target domain for domain adaptation in each round. For V $\rightarrow$ O scenarios, we choose 60 images from target domain randomly.

%---------------------------    Augmentation ablation exp  -----------------------------%
\begin{table*}[!htbp]
  \caption{Instantiations of different types of multi-level domain-aware augmentation. "Source" and "Target" represent that model is only trained in the source and target dataset respectively. "Oracle" means pre-training in the source domain and fine-tuning with the full amount of target domain data. "Spl" and "Rea" denote the image-level domain-splice and domain-reallocation. "Dir", "Gau" and "Mix" represent the direct, Gaussian and mixture box-level exchange methods.}
  \centering
  \renewcommand\tabcolsep{4.0pt} % 调整表格列间的宽度
  \resizebox{\textwidth}{28mm}{
  \begin{tabular}{cccccccccccccc} % 14 col
    \toprule
     & \multicolumn{2}{c}{Img-level} & \multicolumn{3}{c}{Box-level} & \multicolumn{2}{c}{S $
     \rightarrow$ C} & \multicolumn{2}{c}{K $\rightarrow$ C} & \multicolumn{2}{c}{V $\rightarrow$ O} & \multicolumn{2}{c}{C $\rightarrow$ F}  \\
     
    \cmidrule(lr){2-3} \cmidrule(lr){4-6} \cmidrule(lr){7-8} \cmidrule(lr){9-10} \cmidrule(lr){11-12} \cmidrule(lr){13-14}
    % \cmidrule(lr){2-12}
    & Spl & Rea & Dir & Gau & Mix & AP50 & AP & AP50 & AP   & AP50 & AP & mAP50 & mAP\\
    \midrule
    Source & \checkmark & \checkmark & & \checkmark & & 49.0 & 26.5 & 47.4 & 22.4 & 30.4 & 13.0 & 21.9 & 12.0   \\
    Target & \checkmark & \checkmark & & \checkmark & & 75.1 & 49.2 & 75.1 & 49.2 & 81.9 & 56.6 & 45.1 & 21.3    \\
    Oracle   & \checkmark & \checkmark & & \checkmark & & 76.9 & 51.8 & 76.1 & 50.1 & 82.2 & 56.9 & 46.3 & 26.4    \\
    % & & & & & & & & & & & &    \\ \checkmark
    \midrule
    \multirow{12}*{AcroFOD} 
    & & & & & & 55.6\tiny$\pm2.6$ & 33.2\tiny$\pm2.4$ & 51.9\tiny$\pm2.9$ & 27.7\tiny$\pm2.4$ & 42.0\tiny$\pm1.2$ & 19.2\tiny$\pm1.3$ & 26.4\tiny$\pm1.0$ & 15.8\tiny$\pm0.9$   \\
    \specialrule{0em}{1pt}{1pt}
    \cline{2-14}
    \specialrule{0em}{1pt}{1pt}
    &\checkmark & & & & & 60.2\tiny$\pm1.8$ & 36.4\tiny$\pm1.9$ & 58.3\tiny$\pm1.9$ & 32.1\tiny$\pm1.3$ & 44.9\tiny$\pm0.7$ & 22.1\tiny$\pm1.2$ & 27.5\tiny$\pm1.5$ & 15.5\tiny$\pm1.3$   \\
    \specialrule{0em}{1pt}{1pt}
    \cline{2-14}
    \specialrule{0em}{1pt}{1pt}
    &\checkmark & & \checkmark & & & 61.4\tiny$\pm3.0$ & 37.1\tiny$\pm2.3$ & 58.6\tiny$\pm1.7$ & 33.6\tiny$\pm2.1$ & 44.8\tiny$\pm0.6$ & 21.3\tiny$\pm0.6$ & 25.6\tiny$\pm1.3$ & 14.9\tiny$\pm1.1$     \\
    \specialrule{0em}{1pt}{1pt}
    \cline{2-14}
    \specialrule{0em}{1pt}{1pt}
    &\checkmark & & & \checkmark & & 60.3\tiny$\pm2.4$ & 36.3\tiny$\pm2.2$ & 58.4\tiny$\pm2.2$ & 33.1\tiny$\pm2.2$ & 44.5\tiny$\pm1.4$ & 21.1\tiny$\pm1.2$ & 25.1\tiny$\pm1.2$ & 14.3\tiny$\pm0.8$   \\
    \specialrule{0em}{1pt}{1pt}
    \cline{2-14}
    \specialrule{0em}{1pt}{1pt}
    &\checkmark & & & & \checkmark & 61.2\tiny$\pm1.9$ & 36.8\tiny$\pm1.4$ & 57.5\tiny$\pm2.3$ & 32.9\tiny$\pm1.8$ & 43.6\tiny$\pm1.2$ & 20.2\tiny$\pm0.8$ & 28.2\tiny$\pm1.0$ & 16.1\tiny$\pm0.9$   \\
    \specialrule{0em}{1pt}{1pt}
    \cline{2-14}
    \specialrule{0em}{1pt}{1pt}
    &\checkmark & \checkmark & & & & 62.3\tiny$\pm2.7$ & 38.0\tiny$\pm2.3$ & 61.4\tiny$\pm3.1$ & 36.2\tiny$\pm2.3$ & 44.8\tiny$\pm1.3$ & 21.7\tiny$\pm0.6$ & \textbf{41.1}\tiny$\pm0.8$ & \textbf{23.2}\tiny$\pm0.7$   \\
    \specialrule{0em}{1pt}{1pt}
    \cline{2-14}
    \specialrule{0em}{1pt}{1pt}
    &\checkmark & \checkmark & \checkmark & & & 62.2\tiny$\pm3.2$ & 37.7\tiny$\pm2.5$ & 61.6\tiny$\pm2.1$ & 35.5\tiny$\pm2.2$ & 44.7\tiny$\pm0.6$ & 21.5\tiny$\pm0.6$ & 38.6\tiny$\pm0.5$ & 21.9\tiny$\pm0.4$    \\
    \specialrule{0em}{1pt}{1pt}
    \cline{2-14}
    \specialrule{0em}{1pt}{1pt}
    &\checkmark & \checkmark & & \checkmark & & \textbf{62.5}\tiny$\pm1.6$ & \textbf{38.1}\tiny$\pm1.8$ & \textbf{62.6}\tiny$\pm2.1$ & \textbf{36.5}\tiny$\pm2.4$ & 43.9\tiny$\pm1.1$ & 20.4\tiny$\pm0.7$ & 38.3\tiny$\pm0.6$ & 21.4\tiny$\pm0.5$    \\
    \specialrule{0em}{1pt}{1pt}
    \cline{2-14}
    \specialrule{0em}{1pt}{1pt}
    &\checkmark & \checkmark & & & \checkmark & 62.3\tiny$\pm2.1$ & 38.0\tiny$\pm1.0$ & 62.1\tiny$\pm1.9$ & 35.8\tiny$\pm2.3$ & \textbf{45.8}\tiny$\pm0.6$ & \textbf{22.5}\tiny$\pm0.4$ & 36.9\tiny$\pm1.0$ & 20.8\tiny$\pm0.7$   \\
    \bottomrule
  \end{tabular}}
  \label{tab:aug_exp}
\end{table*}
%------------------------------------------------------------------------------------------%
%---------------------------  Sample stategy  exp  -----------------------------%
\begin{table}[t] %!htbp
  \caption{Influence of different strategies for sample selection in S $\rightarrow$ C and K $\rightarrow$ C adaptation scenario.}
  \centering
    \renewcommand\tabcolsep{3.0pt}  % 调整表格列间的长度
  \scalebox{0.9}{\begin{tabular}{ccccc}
    \toprule
    \multirow{2}*{Method}
    & \multicolumn{2}{c}{S $\rightarrow$ C} & \multicolumn{2}{c}{K $\rightarrow$ C} \\
    \cmidrule(lr){2-3} \cmidrule(lr){4-5}
    & AP50 & AP & AP50 & AP  \\
    \midrule
    Source          & 49.0              & 26.5              & 47.4              & 22.4              \\
    \hline
    %Proportion      & 52.5\tiny$\pm1.9$ & 28.9\tiny$\pm1.7$ & 49.8\tiny$\pm1.4$ & 24.2\tiny$\pm1.6$  \\
    Cosine distance & 55.2\tiny$\pm2.7$ & 32.7\tiny$\pm2.6$ & 51.4\tiny$\pm2.1$ & 27.3\tiny$\pm2.3$  \\
    MMD distance    & \textbf{55.6}\tiny$\pm2.6$ & \textbf{33.2}\tiny$\pm2.4$ & \textbf{51.9}\tiny$\pm2.9$ & \textbf{27.7}\tiny$\pm1.8$  \\
    \bottomrule
  \end{tabular}}
  \label{tab:abl_sample}
\end{table}
%------------------------------------------------------------------------------------------%
\noindent\textbf{Instantiations of Domain-aware Augmentation.} Table \ref{tab:aug_exp} shows the effects of different types of domain-aware augmentation in our AcroFOD. For a fair comparison, we choose Eq. \ref{eq:mmd_dis} as distance function in the optimization strategy for all the experiments. From the results, we can notice that either the introduced image-level or the box-level augmentations can both bring significant performance improvements. Meanwhile, combining different types of multi-level domain-aware augmentation can further improve the detection results.  

\noindent\textbf{Different Choices of $dist_f$.} Table \ref{tab:abl_sample} compares different types of distance metric function $dist_f$. The simple proportional sampling strategy achieves better performance than the baseline method trained with source-only data. Optimizing augmented data with MMD distance obtains the best results among all choices. In summary, MMD can better reflect the sample distance between source and target domains than others.

\begin{figure}[ht]
  \centering
  \includegraphics[width=0.98\linewidth]{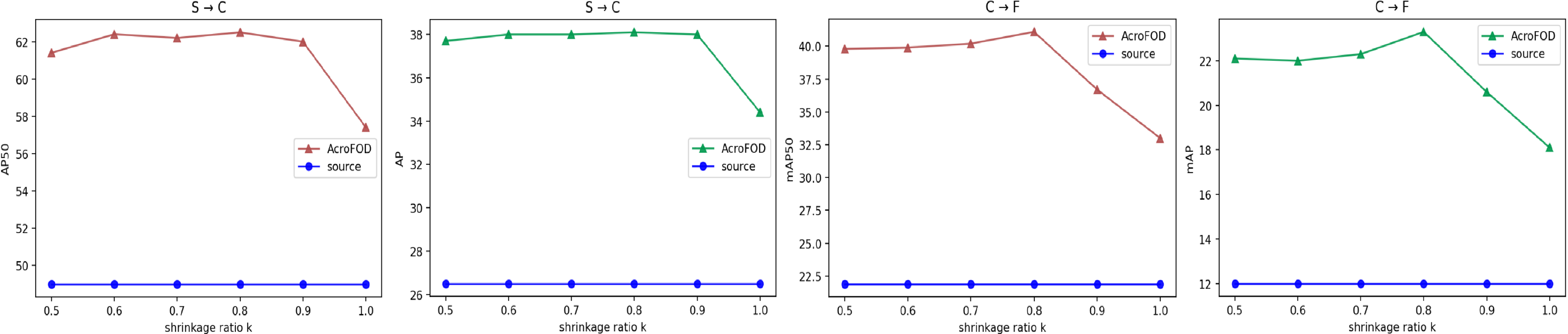}
  \caption{Experiments on different values of $k$ on S $\rightarrow$ C and C $\rightarrow$ F scenarios.}
   \label{fig:sc_k}
\end{figure}

\noindent\textbf{Analysis of Shrinkage Ratio $k$.} Fig. \ref{fig:sc_k} shows the effect of $k$ on S $\rightarrow$ C and C $\rightarrow$ F scenarios. $k = 1$ means training without any process of optimization. All experiments use Eq. \ref{eq:mmd_dis} as $dist_f$. The above two figures show that our proposed adaptive optimization strategy helps model transfer better. Meanwhile, within a certain range of $k$, the performance of the model is relatively stable. 

%---------------------------------ablation study for k-------------------------------------%
%------------------------------------------------------------------------------------------%

\subsection{Experiment with More Data}
\label{sec:more_data}
We conduct a series of experiments to verify the effect of different numbers of data in target domain and choose the challenging V $\rightarrow$ O scenario for verification. We choose 0.1\%, 1\% and 10\% data from COCO person for model training, corresponding to 60, 600 and 6000 images respectively, and test on the validation set of COCO person. The evaluations are conducted on person category and test time on 4 Nvidia V100 GPUs. As shown in Table \ref{tab:coco_more_data}, we notice that the AcroFOD can outperform the general Pre+FT by a large margin in all cases. Meanwhile, our method is faster than both previous work\cite{ramamonjison2021simrod} and Pre+FT paradigm. 
%All these results show that our AcroFOD is more suitable for domain shifts in cross-domain few-shot object detection.

%---------------------------  More data V to O exp  -----------------------------%
\begin{table}[!htbp]
  \caption{Results of more data in V $\rightarrow$ O adaptation scenario. "Source" and "Target" represent that the model is only trained in source and target datasets respectively. "Epochs" means the total number of epochs. "Times" corresponds to the time required to achieve optimum performance.}
  \centering
    \renewcommand\tabcolsep{7.0pt}  % 调整表格列间的长度
  \scalebox{1.0}{
  \begin{tabular}{c|ccc|cc}
    \toprule
    Proportion & Method & AP50     & AP & Epochs & Times  \\
    \midrule
    - & Source & 30.4 & 13.0 & 300 & 48h   \\
    \hline
    \multirow{3}*{$0.1\%$}
    & Target & -    & -  & - & -\\
    %& SimRoD\cite{ramamonjison2021simrod} & 42.8\tiny$\pm1.0$ & 19.5\tiny$\pm0.7$ & 800 & 96h \\ 
    & Pre+FT & 43.2\tiny$\pm1.3$ & 21.0\tiny$\pm0.9$ & 500 &56h  \\     
    & AcroFOD  & \textbf{45.8}\tiny$\pm0.6$ & \textbf{22.5}\tiny$\pm0.4$ & 300 & \textbf{32h}   \\
    \hline
    \multirow{3}*{$1\%$}
    %& Source & 30.4 & 13.0 & 300 & 48h   \\    
    & Target & 43.9\tiny$\pm1.7$ & 20.7\tiny$\pm1.1$ & 300 & 24h   \\
    %& SimRoD\cite{ramamonjison2021simrod} & 46.5\tiny$\pm0.7$ & 23.1\tiny$\pm0.3$ & 800 & 102h \\
    & Pre+FT & 50.9\tiny$\pm0.3$ & 28.2\tiny$\pm0.5$ & 600 & 72h   \\     
    & AcroFOD &  \textbf{61.1\tiny$\pm0.9$} & \textbf{34.1}\tiny$\pm0.6$ & 300 & \textbf{34h}   \\
    \hline
    \multirow{3}*{$10\%$}
    %& Source & 30.4 & 13.0 & 300 & 48h   \\    
    & Target & 71.5\tiny$\pm0.5$ & 44.3\tiny$\pm0.4$ & 300 & 42h   \\
    %& SimRoD\cite{ramamonjison2021simrod} & 67.3\tiny$\pm1.4$ & 43.5\tiny$\pm0.5$ & 800 & 114h \\
    & Pre+FT & 69.7\tiny$\pm0.3$ & 44.5\tiny$\pm0.3$ & 600 & 90h   \\     
    & AcroFOD  & \textbf{75.1}\tiny$\pm0.6$ & \textbf{50.2}\tiny$\pm0.4$ & 300 & \textbf{38h}  \\
    \bottomrule
  \end{tabular} }
  \label{tab:coco_more_data}
\end{table}
%------------------------------------------------------------------------------------------%
%------------------------------------------------------------------------
\section{Conclusion}
We present AcroFOD for adapting detector under domain shift with only a few data in the target domain. Our adaptive optimization for directive data augmentation helps expand limited target data to cover the data distribution of the target domain. Our method achieves significant gains in terms of model robustness compared to existing baselines in few-shot domain adaptation setting. 
The results indicate that the AcroFOD can mitigate the effect of domain shifts due to various changes.
%Our method even obtains comparable performance to unsupervised methods with $300\times$ less than data. 
Through the ablation study, we find some insights on how adaptive optimization and data augmentation from a cross-domain perspective can help model perform better.
We hope this adaptive method will benefit future progress of robust object detection in cross-domain few-shot object detection research.

\noindent\textbf{Acknowledgment.} This work was supported partially by the NSFC (U21A204-71, U1911401, U1811461), Guangdong NSF Project (No.2022A1515011254, 2020-B1515120085, 2018B030312002), Guangzhou Research Project (201902010037), and the Key-Area Research and Development Program of Guangzhou (20200703-0004). 
\bibliographystyle{splncs04}
\bibliography{egbib}
\end{document}